\documentclass[sort&compress]{llncs}
\pdfoutput=1
\usepackage{times}
\usepackage{epsfig}
\usepackage{graphicx}
\usepackage{amsmath,amssymb}
\usepackage{booktabs}
\usepackage{tabulary}
\newcolumntype{C}[1]{>{\centering\arraybackslash}p{#1}}
\usepackage{multirow}
\usepackage{url}
\usepackage{footnote}
\usepackage{tablefootnote}
\usepackage[10pt]{moresize}
\usepackage[hang]{footmisc}
\usepackage{pifont}
\usepackage{lineno,hyperref}
\usepackage{algorithm}
\usepackage{algorithmic}
\usepackage{colortbl}
\usepackage{diagbox}
\usepackage{float}
\usepackage{subfig}
\usepackage{picins}
\usepackage{stfloats}
\usepackage{color}
\usepackage[width=122mm,left=12mm,paperwidth=146mm,height=193mm,top=12mm,paperheight=217mm]{geometry}

\newcommand{\PROC}{\item[\algorithmicproc]}

\newcommand{\algorithmicproc}{\textbf{Procedure:}}

\newcommand{\ktb}[1]{\cellcolor[rgb]{.7843,.8510,.9686}{#1}}

\begin{document}
\pagestyle{headings}
\mainmatter

\def\httilde{\mbox{\tt\raisebox{-.5ex}{\symbol{126}}}}

\newcommand{\specialcell}[2][c]{%
  \begin{tabular}[#1]{@{}l@{}}#2\end{tabular}}

\title{Generalized Haar Filter based Deep Networks for Real-Time Object Detection in Traffic Scene}

\author{Keyu Lu$^1$, Jian Li$^1$, Xiangjing An$^1$, Hangen He$^1$}
\institute{$^1$College of Mechatronic Engineering and Automation, National University of Defense Technology, Changsha 410073, Hunan, China\\
{\tt\small Email: keyu.lu@nudt.edu.cn; keyu.lu@alumni.ubc.ca}}

\maketitle

\begin{abstract}
Vision-based object detection is one of the fundamental functions in numerous traffic scene applications such as self-driving vehicle systems and advance driver assistance systems (ADAS). However, it is also a challenging task due to the diversity of traffic scene and the storage, power and computing source limitations of the platforms for traffic scene applications. This paper presents a generalized Haar filter based deep network which is suitable for the object detection tasks in traffic scene. In this approach, we first decompose a object detection task into several easier local regression tasks. Then, we handle the local regression tasks by using several tiny deep networks which simultaneously output the bounding boxes, categories and confidence scores of detected objects. To reduce the consumption of storage and computing resources, the weights of the deep networks are constrained to the form of generalized Haar filter in training phase. Additionally, we introduce the strategy of sparse windows generation to improve the efficiency of the algorithm. Finally, we perform several experiments to validate the performance of our proposed approach. Experimental results demonstrate that the proposed approach is both efficient and effective in traffic scene compared with the state-of-the-art.
\end{abstract}

\section{Introduction}
\label{intro}

Recent advances in self-driving vehicles and advance driver assistance system (ADAS) have attracted keen attention and interest from researchers and automobile manufacturers. Vision sensor plays an important role in these areas due to its faster response, lower price and power consumption compared with other popular sensors such as LiDAR and millimeter-wave radar. Moreover, vision sensor has the ability to capture rich information from traffic scene (such as luminance, color and texture)~\cite{lky15}, which is beneficial for object detection.

Vision-based object detection is one of the fundamental functions for self-driving vehicle systems and advance driver assistance systems (ADAS), which need to detect the objects around and check whether they are dangerous to the host vehicle. However, object detection in traffic scene from image are still challenging job. On one hand, traffic scenes are diverse and the presence of objects (e.g. vehicles and pedestrians) in traffic scenes are extremely flexible and unpredictable. on the one hand, most traffic scene applications have several special requirements such as real-time, portable (e.g. for ADAS device), low price and power consumption, which are quite different from object detection tasks in the ILSVRC~\cite{ILSVRC} and COCO~\cite{COCO} competitions.

Recently, a great number of researchers have been interested in learning based approaches. Especially in recent years, the success of deep learning boosts the development of vision-based object detection. Convolutional Neural Network (CNN)~\cite{CNN12} is one of the most popular forms of deep networks. By using the strategies of local receptive fields, weight sharing and spatial pooling~\cite{Wang12}, CNN has made a breakthrough in computer vision and image processing. For instance, AlexNet~\cite{CNN12}, which is a type of CNN, has made a startling achievement in the competition of ILSVRC-2012~\cite{ILSVRC} and demonstrated its superiority in image classification. However, with the increasing of network depth, CNN has met the bottleneck in training~\cite{ResNet15}. Recently, He \textit{et al.} proposed the deep residual network~\cite{ResNet15} which employs shortcut connections to overcome the limitations in training a deeper CNNs.

Despite the fact that CNN has achieve a tremendous success in computer vision and image processing, there still exist two main problems that have to be solved when applied to objects detection tasks. Firstly, CNN has a multitude of convolutions that have to be calculated, it would be rather inefficient in object detection if using traditional dense sliding windows paradigm~\cite{Viola04,DPM10}. Secondly, CNN has the property of shift invariance, that is, it is less sensitive to the shift of objects in input image patch. Consequently, it can not achieve precise localization if it is directly applied to objects detection~\cite{Fast_RCNN15}.

To overcome these limitations of CNN, Girshick \textit{et al.} introduced the framework of region proposal based CNN (called R-CNN)~\cite{RCNN12} and successfully applied it to object detection. The main idea of region proposal based CNN is performing CNNs on candidate bounding-boxes (called ``proposals'') which have potential to contain objects~\cite{RCNN12}. Later, the updated version ``Fast R-CNN''~\cite{Fast_RCNN15} is proposed to improve the efficiency of object detection. This method combines R-CNN with SPPnet~\cite{SPPNET14}. In this way, computation of proposals can be sharing and runtime of object detection can be reduced dramatically. However, in both R-CNN and Fast R-CNN, the proposals are generated by Selective Search~\cite{SelectiveSearch13}, which is quite inefficient and thus limits the detection speed of these methods. To overcome this limitation, Faster R-CNN~\cite{FasterRCNN16} proposed to used Region Proposal Networks (RPNs) to generate proposals instead of Selective Search~\cite{SelectiveSearch13}. By this means, the stage proposal generation and CNN-based classification/regression can be performed under an unified framework and thus the detection speed can be boosted with the help of GPU.

However, region proposal based CNNs are complex and not easy to optimize~\cite{YOLO15}. For this reason, several works attempt to achieve real-time objects detection by regarding it as a regression problem. YOLO~\cite{YOLO15} is one of the pioneering works on deep regression networks based object detection. The approach is able to simultaneously outputs the location of bounding box, category and its confidence score for each object in the image at a extremely high frame rate. Nevertheless, it does not work well in small objects detection. In the same vein, Liu \textit{et al.} proposed SSD (Single Shot MultiBox Detector)~\cite{SSD15} for real-time objects detection. It equipped the deep regression networks with several new techniques such as multi-scale feature maps and default boxes~\cite{SSD15}. Despite the fact that it is able to be faster and more accurate compared with YOLO~\cite{YOLO15}, its performance on small objects detection is still unsatisfactory.

\begin{figure*}[t]
	\centering
	\includegraphics[width=1\textwidth,keepaspectratio]{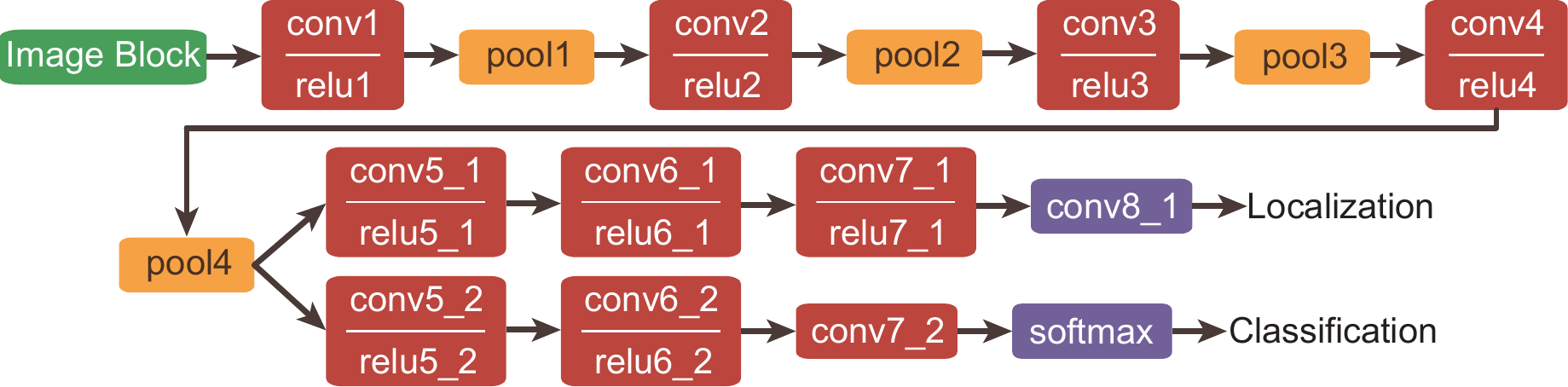}
	\caption{Architecture of the Deep Networks.}
	\label{fig_Pipeline_Net}
\end{figure*}

In this paper, we introduce a practical and robust approach for real-time object detection in traffic scene. The approach requires less storage and computing resources, and thus it is feasible for traffic scene applications. The novelties of this work lie in four fold:

\begin{enumerate}
	\item[(1)]{We present a local regression strategy for accurate objects detection. As we know, current regression based solutions (such as YOLO~\cite{YOLO15} and SSD~\cite{SSD15}) employ global regression strategy to regress the bounding-box of each object from the whole image, which is actually a much more difficult task compared with local regression. Thus, global regression strategy often needs to be supported by a complex or large-scale network and yet its performance on small objects detection is still unsatisfactory. To solve these problem, we introduce the local regression strategy that performs tiny regression networks on small image patches to detect and locate objects. As the regression networks are less sensitive to the scaling and shift of the objects in image patches, a series of sparse sliding-windows can be obtained from multi-scale image pyramid. Finally, objects can be detected and located efficiently based on these sparse sliding-windows;}\vspace{5pt}
	
	\item[(2)]{We introduce the generalized Haar filter based deep networks where each weight larger than \(3 \times 3\) are constrained to the form of generalized Haar filter in training phase. Owing to the strong representation of Haar filters, the networks are able to achieve a high performance. Besides, the networks consume much less storage and computing resources compared with traditional deep networks, which make them possible to be utilized in traffic scene applications. In addition, the constraint of generalized Haar filter provides a form of regularization which is able to improve the generalization ability of the deep networks;}\vspace{5pt}
	
	\item[(3)]{For object detection problems in traffic scene, a sparse windows generation method is proposed. The method first generates a series of sparse sliding-windows in multi-scale image pyramid by setting a specific stride according to the scale and shift tolerance of our deep network. In this way, it can be ensure that each object is completely contained in at least one window. Besides, in most object detection systems for traffic scene, camera is mounted on a fixed position (e.g. mounted on the top of a vehicle windshield). Consequently, objects (e.g. vehicles and pedestrians) produce the predetermined location-specific patterns in images. According to this assumption, perspective geometry is also utilized to further reduce the candidate windows;}\vspace{5pt}
	
	\item[(3)]{We construct a tiny deep network that simultaneously outputs the bounding box, category and confidence score of detected object through two output channels: localization channel and classification channel. The network is efficient and consumes less resources. As our proposed method decomposes the global regression task into several easier local regression tasks, which can be handled effectively without the support of the complex or large-scale networks. Thus, our proposed approach can work efficiently and effectively by combining this tiny a deep network with local regression tasks.}	
\end{enumerate}

The rest of this paper is organized as follows: In section~\ref{rec_GHaar}, we introduce the architecture of our generalized Haar filter based deep networks and describe how to design their weights. In section~\ref{sec_sparse_windows}, we propose the algorithm of sparse windows generation. Experimental results and corresponding discussion are presented in section~\ref{sec_res_dis}, while conclusions and future works are given in section~\ref{sec_conclu}.

\section{Generalized Haar Filter based Deep Networks}
\label{rec_GHaar}

\subsection{Architecture of the Deep Networks}

\begin{figure}[t]
	\centering
	\includegraphics[width=0.2\textwidth,keepaspectratio]{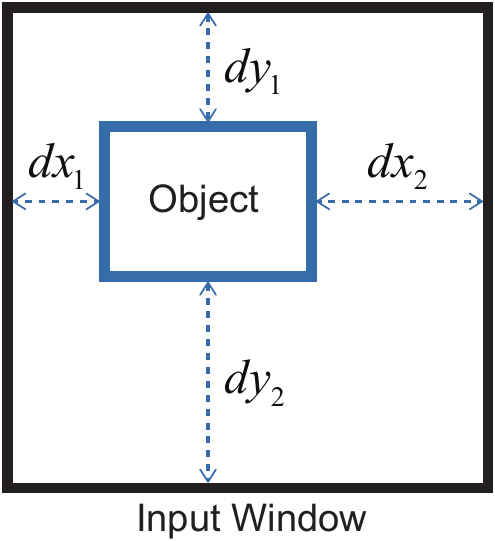}
	\caption{Description of a bounding box using a 4-dimensional vector.}
	\label{fig_LocShow}
\end{figure}

To solve the problem of object detection, we construct a deep network that simultaneously outputs the bounding box, category and confidence score of detected object via two output channels: localization channel and classification channel (see Fig.~\ref{fig_Pipeline_Net}). The location channel focuses on bounding box regression and outputs a 4-dimensional vector \(( d{x_1},d{x_2},d{y_1},d{y_2}) \) which is used to describe a bounding box (see Fig.~\ref{fig_LocShow}). The classification channel outputs the category and confidence score which are denoted by a 2-dimensional vector \(( l,s ) \).

As shown in Fig.~\ref{fig_Pipeline_Net}, our deep network consists of 11 convolution layers, 4 max-pooling layers and 1 softmax layer. To reduce the memory consumption, several low-level features are shared by localization and classification channels via conv1$\sim$pool4. As these two channels aim at different tasks, each of them has 4 independent layers to focus on different problems.

Our deep network is less sensitive to the scaling and shift of the input object owing to regression based localization channel. Consequently, instead of generating region proposals, objects can be detected and located efficiently based on sparse sliding-window paradigm and perspective geometry. More details on this issue are introduced in section~\ref{subsec_sliding_windows} and related experiments are presented in section~\ref{sec_res_dis}.

\subsection{Generalized Haar Filter based Weights Design}

Despite the fact that deep neural networks have achieved state-of-the-art performance in object detection, they consume considerable storage, computing resources and power~\cite{DeepCompress}. Consequently, they are often unsuitable for power and memory constrained devices such as vehicle-mounted embedded systems and mobile devices. To overcome this limitation, we introduce a novel convolution weights design method which is based on generalized Haar filter.

Haar-like filters have been successfully applied in object detection owing to their strong representation and high efficiency~\cite{Viola04,CompressTracking12}. Instead of using a fixed number of rectangles and configuration types, generalized Haar filters are based on arbitrary configurations and number of rectangles~\cite{GHaarLike05}. In this work, the weights of our deep network are constrained based on generalized Haar filters.

Unlike original generalized Haar filters with arbitrary configuration and number of rectangles~\cite{GHaarLike05}, in our work, the configuration and number of rectangles are obtained in a data-driven way. For a weight \({w_i}\) of size \(m \times m\) (\(m \ge 3\)), we constrain it to the following form:
\begin{equation}
{w_i} = {{\hat w}_p} \cdot {k_i},
\end{equation}
where \({k_i} \in R\) is a multiplication factor and \({{\hat w}_p}\) is the \(p\)-th generalized Haar filter in the Haar filters space. As we know, there are \({2^{{m^2}}}\) types of configurations for the generalized Haar filters of size \(m \times m\). In our case, \({{\hat w}_p}\) and its negative form \( - {{\hat w}_p}\) can be regarded as a same filter. Thus, the Haar filter space of size \(m \times m\) contains \({2^{{m^2} - 1}}\) filters. That is, \(p \in [1,{2^{{m^2} - 1}}]\).

\begin{figure}[t]
	\centering
	\includegraphics[width=0.47\textwidth,keepaspectratio]{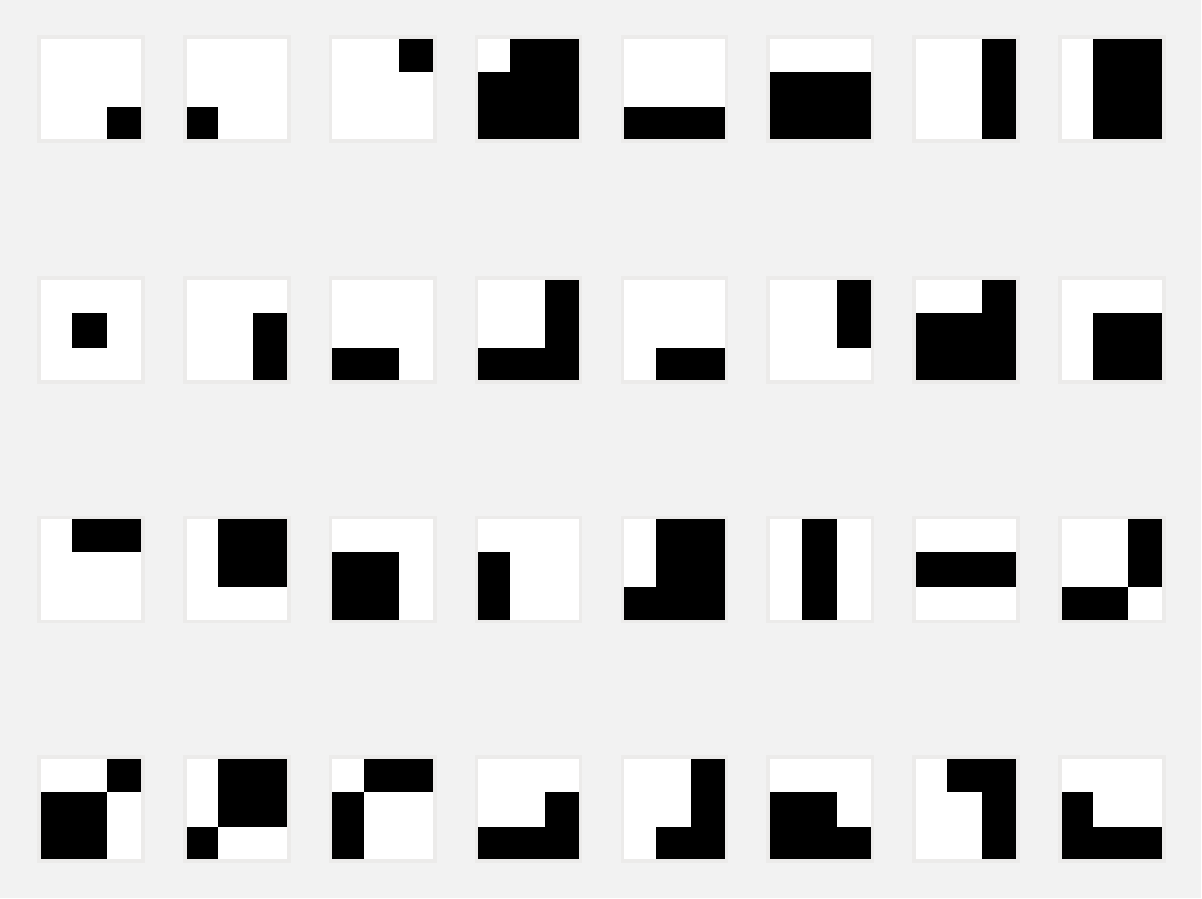}
	\caption{The illustration of the selected generalized Haar filters.}
	\label{fig_HaarSel}
\end{figure}

\begin{figure*}[t]
	\centering
	\includegraphics[width=1\textwidth,keepaspectratio]{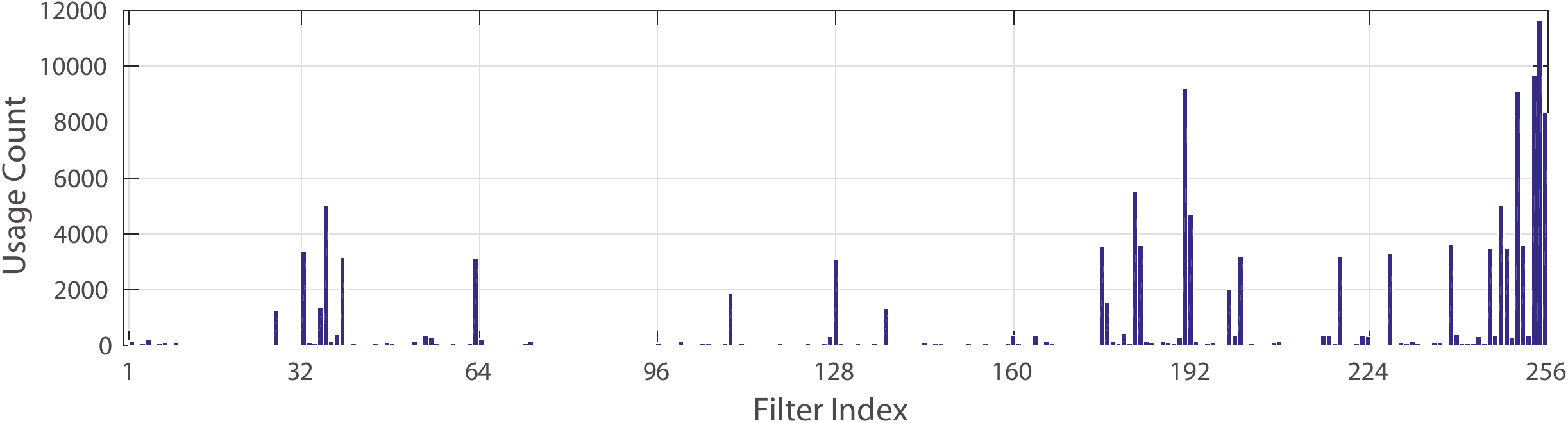}
	\caption{Filter usage in \(3 \times 3\) Haar filter space.}
	\label{fig_UsageCount}
\end{figure*}

For a trained deep network for vehicle detection, Fig. (\ref{fig_UsageCount}) illustrates the filter usage in the \(3 \times 3\) Haar filter space. As shown in Fig. (\ref{fig_UsageCount}), filter usage is quite ``sparse''. In other words, a few generalized Haar filters tend to be used much more frequent than the rest of filters in \(3 \times 3\) Haar filter space. Thus, these filters are more important and representative than the other filters. According to this fact, we try to reserve these ``important'' generalized Haar filters and remove the rest of filters from Haar filter space. In this way, the configuration and number of rectangles of generalized Haar filters are obtained in a data-driven way. Let \(Nr\) denote the number of filter we try to reserve. We sort the filters according to their usage count and select the top \(Nr\) filters (\(Nr = 32\) in our work). These selected filters are shown in Fig. (\ref{fig_HaarSel}). Then, the deep network is retrained by constraining the corresponding weights to these \(Nr\) filters with multiplication factors.

Generalized Haar filters are able to compress the deep networks by locking the relationship between each element in the weights larger than \(3 \times 3\). In this way, considerable storage resources can be saved. For each weight of our deep network, we only need to store a multiplication factor and a filter index. In this work, each multiplication factor (single-precision float-point) takes 4 bytes and each filter index (ranging from 1 to 32) takes less than 1 byte. Thus, only 5 bytes are needed for each weight which is larger than \(3 \times 3\).  By contrast, in traditional deep networks, \(4{m^2}\) bytes are consumed for each weight of size \(m \times m\). Besides, despite the fact that weights are constrained according to \(Nr\) generalized Haar filters, each element of the weights is still able to keep a relatively high precision owing to the multiplication factors. 

Unlike approaches in~\cite{CompressTracking12,Viola04,GHaarLike05} where Haar filters are operated based on integral images. We calculate Haar filter based convolution by basic addition and multiplication operation. One reason is that most weights in deep networks are relatively small (\(3 \times 3\) and \(5 \times 5\)), accordingly, there is no obvious benefit in using integral images. Moreover, integral image of each channel is needed to be re-computed for each layer, which is inefficient. Besides, the configuration types and number of rectangles of the generalized Haar filter are diverse due to our data-driven filter selection strategy, consequently, lots of different computation rules are needed to be designed for these selected filters if calculated based on integral images. In this way, it difficult to achieve efficient batch calculation.

In our work, generalized Haar filters only contains two types of elements: -1 and +1. Thus, each generalized Haar filter can be regarded as a sign pattern matrix. Therefore, the weight \({w_i}\) can be written as:
\begin{equation}
{w_i} = {{\hat w}_p} \cdot {k_i} = {\rm{sign}}({{\hat w}_p}) \cdot k.
\end{equation}

As we know, each convolution step can be regarded as a dot product operation. Let \({P_i}\) denote the dot product between the Haar filter \({{\hat w}_p}\) and the input patch \({x_i}\), that is:
\begin{equation}
{P_i} = {{\hat w}_p} \cdot {x_i} = {\rm{sign}}({{\hat w}_p}) \cdot {x_i}.
\label{eq_dot_prod}
\end{equation}

As a Haar filter can be regarded as a sign pattern matrix, equation (\ref{eq_dot_prod}) can be calculated by lookup table without multiplication. Then each convolution step can be transformed to the following form:
\begin{equation}
{w_i} \cdot {x_i} = ({{\hat w}_p} \cdot {x_i}) \cdot {k_i} = {P_i} \cdot {k_i} = {k_i} \cdot \sum {{P_i}},
\label{eq_convstep}
\end{equation}
where ``\(\sum \)'' denotes the sum of the matrix elements. In this way, only one multiplication is needed for each convolution step. By contrast, in traditional deep networks, \({m^2}\) multiplication are needed. In a work, our approach consume much less computing resources, in this way, power consumption can be also reduced. Accordingly, our deep network is suitable for embedded devices and FPGA where power and available multipliers are limited.

In addition, the constraint of generalized Haar filter provides a form of regularization which is able to improve the generalization ability of the deep network. The loss function and the corresponding regularization term are introduced in the next subsection.

\subsection{Multi-Task Training}

Our deep network has two output channels: localization channel and classification channel. Each localization channel outputs a 4-dimensional vector \(d = (d{x_1},d{x_2},d{y_1},d{y_2})\). Given the ground truth location vector \(\hat d = (d{{\hat x}_1},d{{\hat x}_2},d{{\hat y}_1},d{{\hat y}_2})\), the localization loss can be defined as a squared loss form:
\begin{equation}
{L_{loc}}(d,\hat d) = {\left\| {d - \hat d} \right\|^2}.
\end{equation}

As classification channel focuses on a typical classification problem, we define the classification loss \({L_{cla}}\) as a traditional softmax-loss form.

For a weight \({{w_i}}\) (\(m \times m\), \(m \ge 3\)) in our deep network, the training goals is to obtain a Haar filter index \(p\) and a multiplication factor \({k_i}\). Each \({{w_i}}\) is constrained according to least squares principle, that is:
\begin{equation}
\min {\left\| {{{({w_i} - {{\hat w}_r} \cdot {\lambda _r})}^2}} \right\|_1}, \quad r = 1,2, \ldots ,{2^{{m^2} - 1}}.
\end{equation}

For each Haar filter in the Haar filter space, the corresponding multiplication factor \({{\lambda _r}}\) can be obtained by:
\begin{equation}
\frac{d}{{d{\lambda _r}}}{\left\| {{{({w_i} - {{\hat w}_r} \cdot {\lambda _r})}^2}} \right\|_1} = 0.
\label{eq_deri}
\end{equation}

From equation (\ref{eq_deri}), \({{\lambda _r}}\) can be calculated:
\begin{equation}
{\lambda _r} = \frac{{\sum {{w_i} \cdot {{\hat w}_r}} }}{{\sum {\hat w_r^2} }}.
\end{equation}

Having obtained \({{\lambda _r}}\), we constrain each weight by adding a regularization term to its original loss function \(C{o_i}({w_i},w_i^*)\):
\begin{equation}
{C_i}({w_i},w_i^*) = C{o_i}({w_i},w_i^*) + \varphi  \cdot \mathop {\min }\limits_r {\left\| {{{({w_i} - {{\hat w}_r} \cdot {\lambda _r})}^2}} \right\|_1}.
\end{equation}

This regularization term can not only reduce the storage consumption of our deep network, but also improve the generalization ability of the deep network. Further experiments are presented in section~\ref{sec_res_dis}.

As the deep network are trained via stochastic gradient descent (SGD), the loss function \({C_i}({w_i},w_i^*)\) is needed to be transformed to a differentiable form. According to~\cite{MinMax11}, a maximum function for \(Z = \{ {z_1}, \ldots ,{z_n}\} \) can be smoothly approximated as:
\begin{equation}
{F_{\max }}(Z) = \frac{{\sum\limits_{r = 1}^n {z_r^{q + 1}} }}{{\sum\limits_{r = 1}^n {z_r^q} }},
\end{equation}
where \(p \ge 1\), \({\sum\limits{z_r^q} } \ne 0\) and \({z_i} \ge 0\). The continuity and differentiability have been proofed by~\cite{MinMax11}. Inspired by this work, we first define the intermediate variable \({z_r}\) as:
\begin{equation}
\begin{split}
\ {z_r} &= \exp ( - {\left\| {{{({w_i} - {{\hat w}_r} \cdot \lambda )}^2}} \right\|_1}) \\
\  &= \exp ( - {\left\| {{{({w_i} - {{\hat w}_r} \cdot \frac{{\sum {{w_i} \cdot {{\hat w}_r}} }}{{\sum {\hat w_r^2} }})}^2}} \right\|_1}).
\end{split}
\end{equation}

Then the loss function of each weight can be transformed to a differentiable form:
\begin{equation}
{C_i}({w_i},w_i^*) = C{o_i}({w_i},w_i^*) - \varphi  \cdot \ln \left( {\frac{{\sum\limits_{r = 1}^{{2^{{m^2} - 1}}} {z_r^{q + 1}} }}{{\sum\limits_{r = 1}^{{2^{{m^2} - 1}}} {z_r^q} }}} \right).
\label{eq_weightloss}
\end{equation}

Having obtained the differentiable form of the loss function of each weight, the updated weight \({w_i^{t + 1}}\) can be obtained using stochastic gradient descent (SGD). Then, the Haar filter index \(p\) and a multiplication factor \({k_i}\) can be obtained by:
\begin{equation}
{
	\left\{
	\begin{aligned}
	\! p &= \mathop {\arg \min }\limits_r {\left\| {{{(w_i^{t + 1} - {{\hat w}_r} \cdot \frac{{\sum {w_i^{t + 1} \cdot {{\hat w}_r}} }}{{\sum {\hat w_r^2} }})}^2}} \right\|_1} \\
	\! {k_i} &= \frac{{\sum {w_i^{t + 1} \cdot {{\hat w}_p}} }}{{\sum {\hat w_p^2} }}
	\end{aligned}
	\right. .
}
\label{eq_haar_update}
\end{equation}

The procedure for training our deep network is demonstrated in algorithm~\ref{alg_training}. For the \(i\)-th weight of our deep network, we first construct the current weight using the filter index and multiplication factor which are updated in previous iteration. Then, the target \(w_i^*\) are obtained via a standard forward propagation approach. After that, based on the loss function \({C_i}(w_i^t,w_i^*)\) (see equation (\ref{eq_weightloss})), the gradients can be computed using a standard backward propagation approach. Having obtained the gradients, weight is updated using stochastic gradient descent (SGD). Lastly, using the updated weight, filter index and multiplication factor can be updated via equation (\ref{eq_haar_update}).

\begin{algorithm}[htb]
	\caption{Parameter update of the \(i\)-th weight.}
	\label{alg_training}
	\begin{algorithmic}[1]
		\REQUIRE \hspace{1.2ex} A minibatch of inputs \({x_i}\); \\
		\hspace{6.5ex}Loss function \({C_i}(w_i^t,w_i^*)\); \\
		\hspace{6.5ex}Generalized Haar filter space \({{\hat w}_r}\), \\ 
		\hspace{6.5ex}(\(r = 1,2, \ldots ,{2^{{m^2} - 1}}\)); \\
		\hspace{6.5ex}Current filter index \({\tilde p}\);\\
		\hspace{6.5ex}Current multiplication factor \({{\tilde k}_i}\); \\
		\hspace{6.5ex}Current learning rate \({l_t}\). \\
		
		\ENSURE Updated filter index \(p\); \\
		\hspace{6.5ex}Updated multiplication factor \({k_i}\); \\
		\hspace{6.5ex}Updated learning rate \({l_{t + 1}}\).\\
		
		\PROC ~\\
		
		\STATE Constructing current weight: \\ \centerline{\(w_i^t = {w_{\tilde p}} \cdot {{\tilde k}_i}\);}
		
		\STATE Obtaining target using standard forward propagation: \\ \centerline{\(w_i^* = {\rm{Forward}}(w_i^t,{x_i})\);}
		
		\STATE Gradients are computed via standard backward propagation: \\ \centerline{\(\frac{{\partial {C_i}}}{{\partial w_i^t}} = {\rm{Backward(}}w_i^t,w_i^*)\);}
		
		\STATE Updating weight using stochastic gradient descent: \\ \centerline{\(w_i^{t + 1}{\rm{ = UpdateWeight}}(w_i^t,\frac{{\partial {C_i}}}{{\partial {w_i}}},{l_t})\);}
		
		\STATE Updating filter index: \\ \centerline{\(p = \mathop {\arg \min }\limits_r {\left\| {{{(w_i^{t + 1} - {{\hat w}_r} \cdot \frac{{\sum {w_i^{t + 1} \cdot {{\hat w}_r}} }}{{\sum {\hat w_r^2} }})}^2}} \right\|_1}\);}
		
		\STATE Updating multiplication factor: \\ \centerline{\({k_i} = \frac{{\sum {w_i^{t + 1} \cdot {{\hat w}_p}} }}{{\sum {\hat w_p^2} }}\);}
		
		\STATE Updating learning rate: \\ \centerline{\({l_{t + 1}}{\rm{ = UpdateLearningRate}}({l_t},t)\).}
		
	\end{algorithmic}
\end{algorithm}

\section{Sparse Windows Generation}
\label{sec_sparse_windows}

Due to the regression based localization channel, our deep network is less sensitive to the scaling and shift of the input object. Consequently, instead of traditional dense sliding-window paradigm, we employ sparse sliding-window strategy to achieve real-time object detection in traffic scene. Besides, in most object detection systems for traffic scene, camera is mounted on a fixed position (e.g. mounted on the top of a vehicle windshield). Accordingly, objects (e.g. vehicles and pedestrians) produce the predetermined location-specific patterns in images. Thus, the potential appearance of objects in the images can be obtained using the perspective geometry of the given scene. Based on the perspective geometry, a series of sparse windows can be generated according to the scale and shift tolerance of our deep network. Finally, the locations and categories of objects in the given image can be obtained efficiently by performing our deep network on these sparse windows.

\subsection{Sparse Sliding-Window Strategy}
\label{subsec_sliding_windows}
\begin{figure}[t]
	\centering
	\includegraphics[width=0.2\textwidth,keepaspectratio]{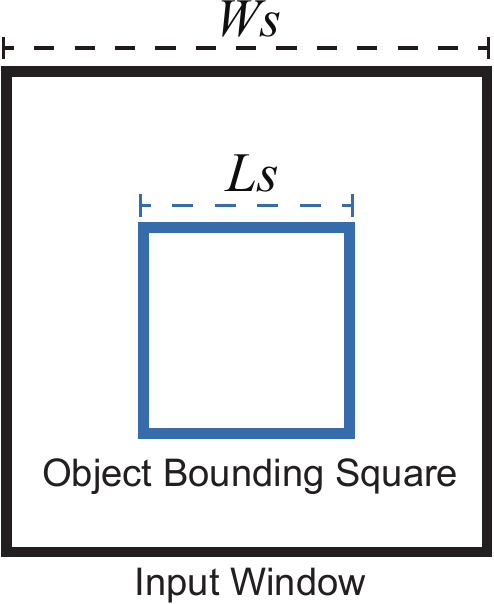}
	\caption{Illustration of object size ratio.}
	\label{fig_size_ratio}
\end{figure}

We first generate a set of input windows \({U_s}\) using sparse sliding-window strategy. In order to obtain \({U_s}\) which is able to fit the objects in different size, we define a size ratio for each input window. As shown in Fig.~\ref{fig_size_ratio}, let \(Ws\) denote the size of each input window and \(Ls\) denote the size of each object bounding square. The size ratio between object bounding square and the corresponding input window is represented by \(Rs\):
\begin{equation}
Rs = \frac{{Ls}}{{Ws}}.
\end{equation}

We then generate image pyramid by resizing the given image to different scales. For each resized image in the image pyramid, we assume that each deep network is responsible for objects with \(Rs\) ranging from 0.5$\sim$0.7 (objects with Rs beyond this range would be detected in other image scale in the image pyramid). By setting the stride of sliding-windows to 0.3, it can be ensure that each object is completely contained in at least one window. Thus, our approach is able to combine the benefits of regression based method and sliding-window based method. Instead of applying global regression on whole image, our approach performs local regression for bounding box localization. Consequently, the approach has the potential to detected smaller objects compared with global regression based method~\cite{YOLO15,SSD15}. Besides, the approach maintains a relatively high efficiency thanks to the sparse sliding-window paradigm. Further experiments are presented in section~\ref{sec_res_dis}.

\subsection{Perspective Geometry}
Having obtained the set of input windows \({U_s}\) by sparse sliding-window strategy. We then generate a set of input windows \({U_p}\) according to the perspective geometry. Let \(({x_{3D}},{y_{3D}},{z_{3D}})\) denote a 3D point position in world coordinates, \(({x_{2D}},{y_{2D}})\) denotes the corresponding 2D point position in pixel coordinates. According to the perspective geometry, they satisfy:
\begin{equation}
z\left[ {\begin{array}{*{20}{c}}
	{{x_{2D}}}\\
	{{y_{2D}}}\\
	1
	\end{array}} \right] = M\left[ {\begin{array}{*{20}{c}}
	{{x_{3D}}}\\
	{{y_{3D}}}\\
	{{z_{3D}}}\\
	1
	\end{array}} \right],
\label{eq_2d3d}
\end{equation}
where \(M\) is a camera projection matrix which is the product of an intrinsic matrix and an extrinsic matrix:
\begin{equation}
M = \left[ {\begin{array}{*{20}{c}}
	{{f_x}}&0&{{u_0}}&0\\
	0&{{f_y}}&{{v_0}}&0\\
	0&0&1&0
	\end{array}} \right]\left[ {\begin{array}{*{20}{c}}
	{\bf{R}}&{\bf{T}}\\
	{{{\bf{0}}^{\bf{T}}}}&1
	\end{array}} \right].
\label{eq_M_MAT}
\end{equation}

In equation (\ref{eq_M_MAT}), \(R\) is a rotation matrix which is the result of three rotations around the world coordinate axes. As in most object detection systems for traffic scene, the angles of camera rotations are relatively small. Accordingly, the camera projection matrix \(M\) can be approximated as:
\begin{equation}
M = \left[ {\begin{array}{*{20}{c}}
	{{m_{11}}}&0&{{m_{13}}}&{{m_{14}}}\\
	0&{{m_{22}}}&{{m_{23}}}&{{m_{24}}}\\
	0&0&1&{{m_{34}}}
	\end{array}} \right].
\end{equation}

Then, the equation (\ref{eq_2d3d}) can be transformed to:
\begin{equation}
{
	\left\{
	\begin{aligned}
	\! {x_{2D}} &= \frac{{{m_{11}} \cdot {x_{3D}} + {m_{13}} \cdot {z_{3D}} + {m_{14}}}}{{{z_{3D}} + {m_{34}}}} \\
	\! {y_{2D}} &= \frac{{{m_{22}} \cdot {y_{3D}} + {m_{23}} \cdot {z_{3D}} + {m_{24}}}}{{{z_{3D}} + {m_{34}}}}
	\end{aligned}
	\right. .
}
\label{eq_fxy}
\end{equation}

In this work, each input window of our deep network has the same height and width. We use \({d_{2D}}\) to represent the height or width of the input window located at \(({x_{2D}},{y_{2D}})\) in pixel coordinates (center aligned),  and let \({d_{3D}}\) denote the corresponding height or width in world coordinates. Then \({d_{2D}}\) can be formulated as:
\begin{equation}
{d_{2D}} = \frac{{{m_{11}} \cdot {d_{3D}}}}{{{z_{3D}} + {m_{34}}}}.
\label{eq_widhei}
\end{equation}

As we know, the locations of most objects in traffic scene are limited (e.g. vehicles and pedestrians will not appear on sky region). In this work, the location and size of each input window of our deep network is described by a triplet \(({x_{2D}},{y_{2D}},{d_{2D}})\) in pixel coordinates. Let \([x_{3D}^{\min },x_{3D}^{\max }]\) and \([y_{3D}^{\min },y_{3D}^{\max }]\) respectively represent the location ranges of the corresponding windows in \({{x_{3D}}}\) and \({{y_{3D}}}\) axes of the world coordinates. By solving the equation set that consists of equation (\ref{eq_fxy}) and (\ref{eq_widhei}), we can find 4 boundary planes in \({x_{2D}}{y_{2D}}{d_{2D}}\) space which jointly limit the locations and sizes of input windows:
\begin{equation}
{
	\left\{
	\begin{aligned}
	\! {d_{2D}} &= {x_{2D}} \cdot {g_x}(x_{3D}^{\min }) - {g_x}(x_{3D}^{\min }) \cdot {m_{13}} \\
	\! {d_{2D}} &= {y_{2D}} \cdot {g_y}(y_{3D}^{\min }) - {g_y}(y_{3D}^{\min }) \cdot {m_{23}} \\
	\! {d_{2D}} &= {x_{2D}} \cdot {g_x}(x_{3D}^{\max }) - {g_x}(x_{3D}^{\max }) \cdot {m_{13}} \\
	\! {d_{2D}} &= {y_{2D}} \cdot {g_y}(y_{3D}^{\max }) - {g_y}(y_{3D}^{\max }) \cdot {m_{23}}
	\end{aligned}
	\right.,
}
\label{eq_eqset}
\end{equation}
where:
\begin{equation}
{
	\left\{
	\begin{aligned}
	\! {g_x}({x_{3D}}) &= \frac{{{m_{11}} \cdot {d_{3D}}}}{{{m_{11}} \cdot {x_{3D}} - {m_{13}} \cdot {m_{34}} + {m_{14}}}} \\
	\! {g_y}({y_{3D}}) &= \frac{{{m_{11}} \cdot {d_{3D}}}}{{{m_{22}} \cdot {y_{3D}} - {m_{23}} \cdot {m_{34}} + {m_{24}}}}
	\end{aligned}
	\right..
}
\end{equation}

\begin{figure}[t]
	\centering
	\includegraphics[width=0.47\textwidth,keepaspectratio]{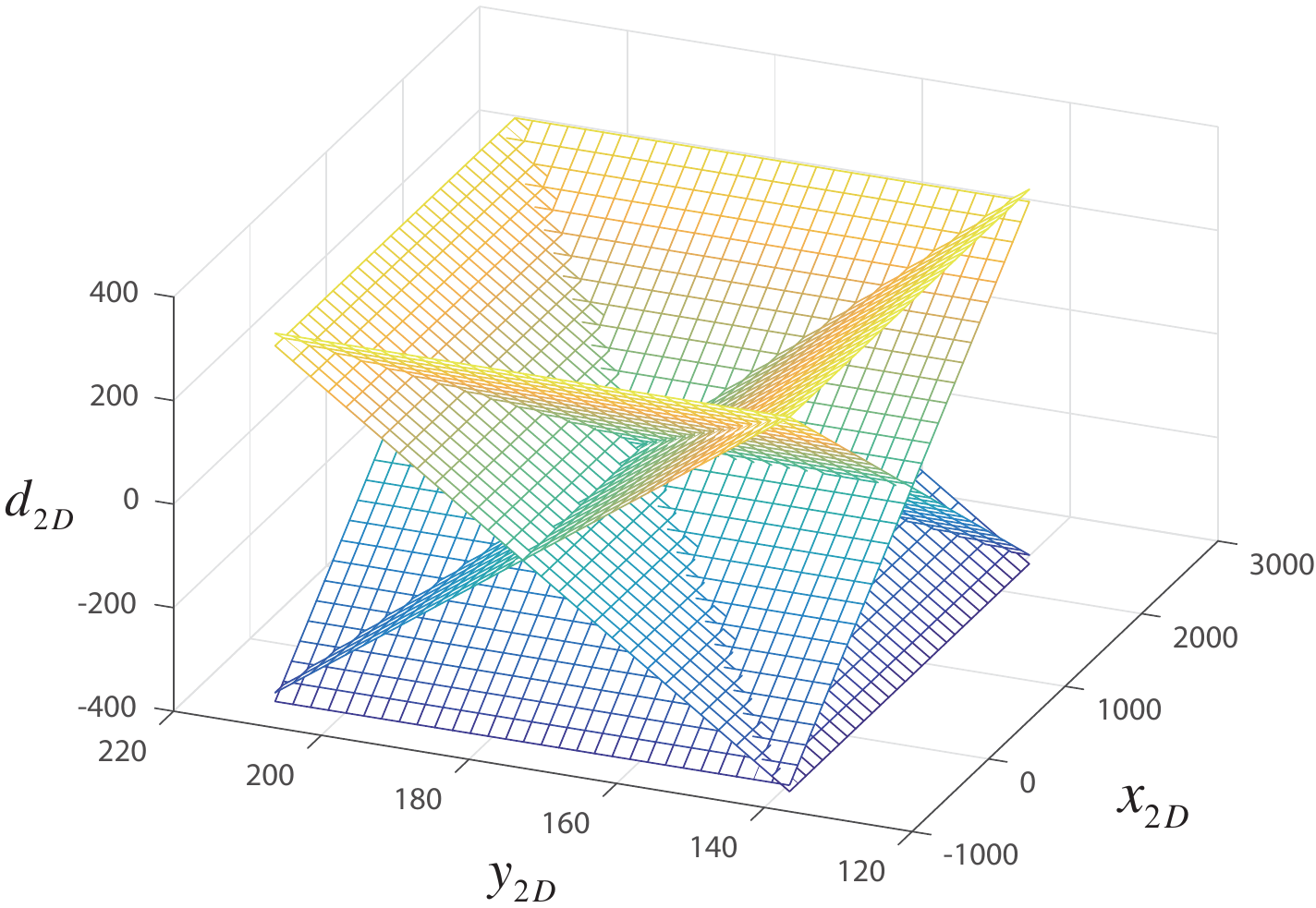}
	\caption{Illustration of 4 boundary planes that jointly limit the locations and sizes of input windows in \({x_{2D}}{y_{2D}}{d_{2D}}\) space.}
	\label{fig_boundaryplane}
\end{figure}

As shown in Fig.~\ref{fig_boundaryplane}, in the \({x_{2D}}{y_{2D}}{d_{2D}}\) space, possible input windows are distributed in the region of inverted pyramid which is enclosed by 4 boundary planes. We use \({U_p}\) to represent the set of input windows in this region, then the final sparse windows \({U_f}\) can be obtained by:
\begin{equation}
{U_f} = {U_p} \cap {U_s}.
\end{equation}

\begin{figure*}[t]
	\centering
	{
		\begin{minipage}[t]{1\linewidth}
			\centering  
			{
				\subfloat[]
				{
					\label{fig_Exp_Py01}
					\begin{minipage}[t]{0.24\linewidth}
						\centerline
						{
							\includegraphics[width=1\textwidth]{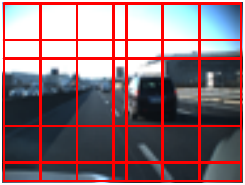}
						}
					\end{minipage}
				}
				\subfloat[]
				{
					\label{fig_Exp_Py02}
					\begin{minipage}[t]{0.24\linewidth}
						\centerline
						{
							\includegraphics[width=1\textwidth]{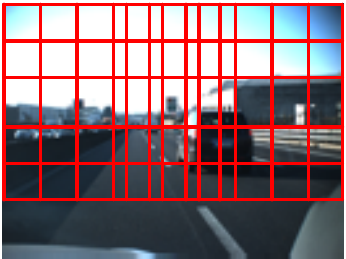}
						}
					\end{minipage}
				}
				\subfloat[]
				{
					\label{fig_Exp_Py03}
					\begin{minipage}[t]{0.24\linewidth}
						\centerline
						{
							\includegraphics[width=1\textwidth]{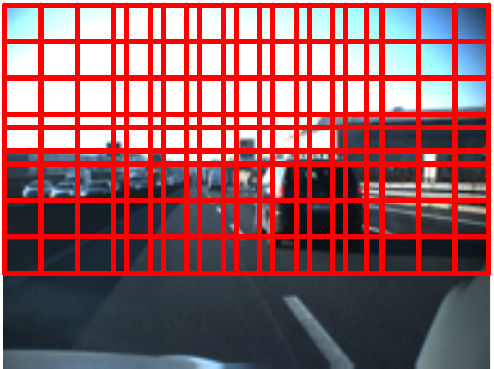}
						}
					\end{minipage}
				}
				\subfloat[]
				{
					\label{fig_Exp_Py04}
					\begin{minipage}[t]{0.24\linewidth}
						\centerline
						{
							\includegraphics[width=1\textwidth]{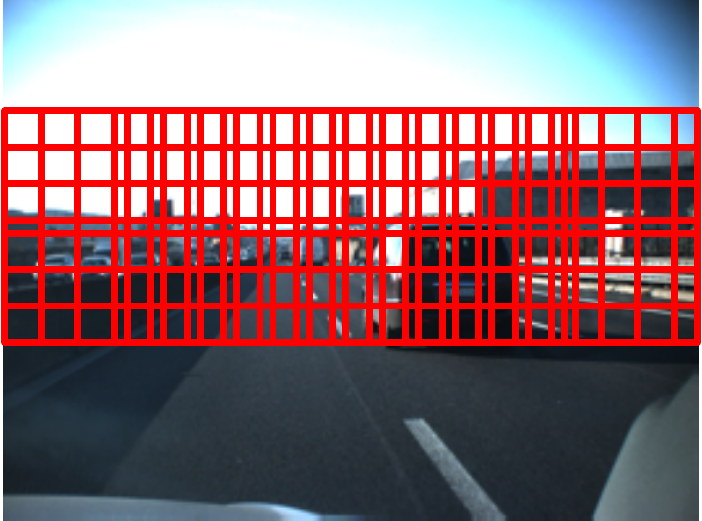}
						}
					\end{minipage}
				}
			}
			\\
			\centering
			{
				\subfloat[]
				{
					\label{fig_Exp_Py05}
					\begin{minipage}[t]{0.24\linewidth}
						\centerline
						{
							\includegraphics[width=1\textwidth]{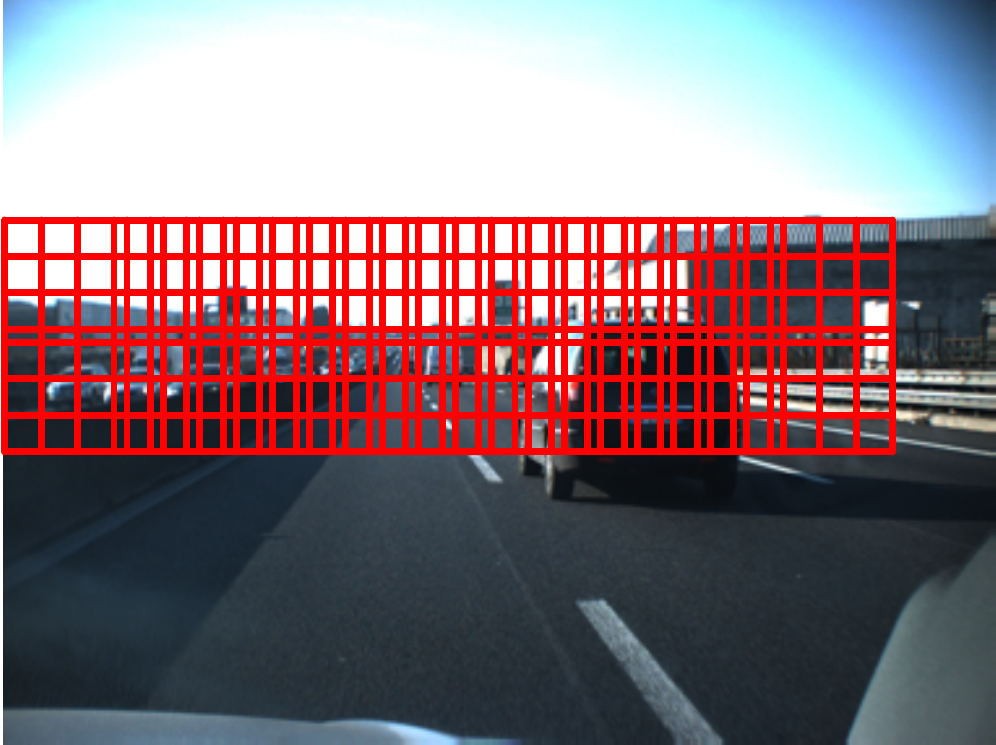}
						}
					\end{minipage}
				}
				\subfloat[]
				{
					\label{fig_Exp_Py06}
					\begin{minipage}[t]{0.24\linewidth}
						\centerline
						{
							\includegraphics[width=1\textwidth]{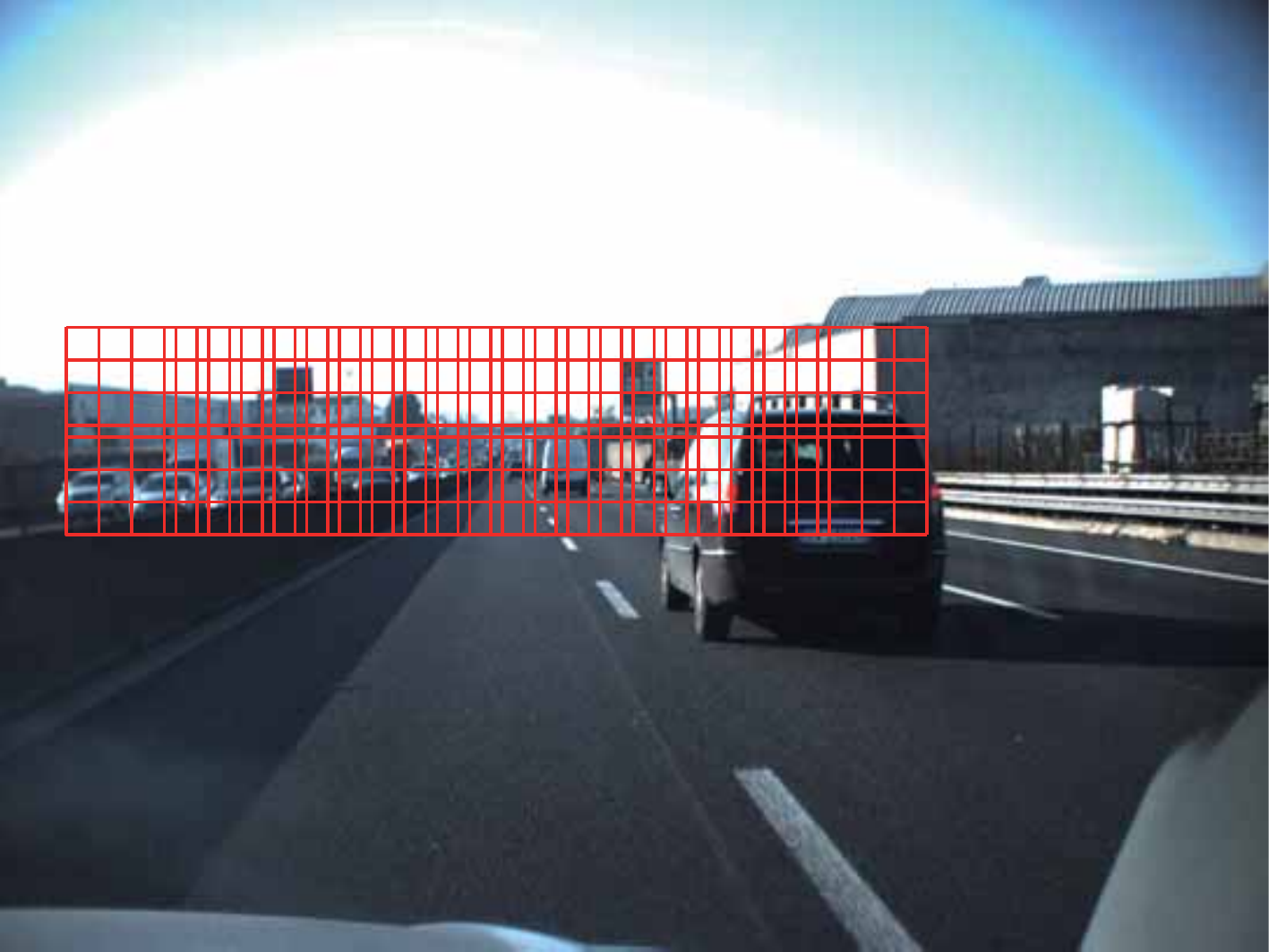}
						}
					\end{minipage}
				}
				\subfloat[]
				{
					\label{}
					\begin{minipage}[t]{0.24\linewidth}
						\centerline
						{
							\includegraphics[width=1\textwidth]{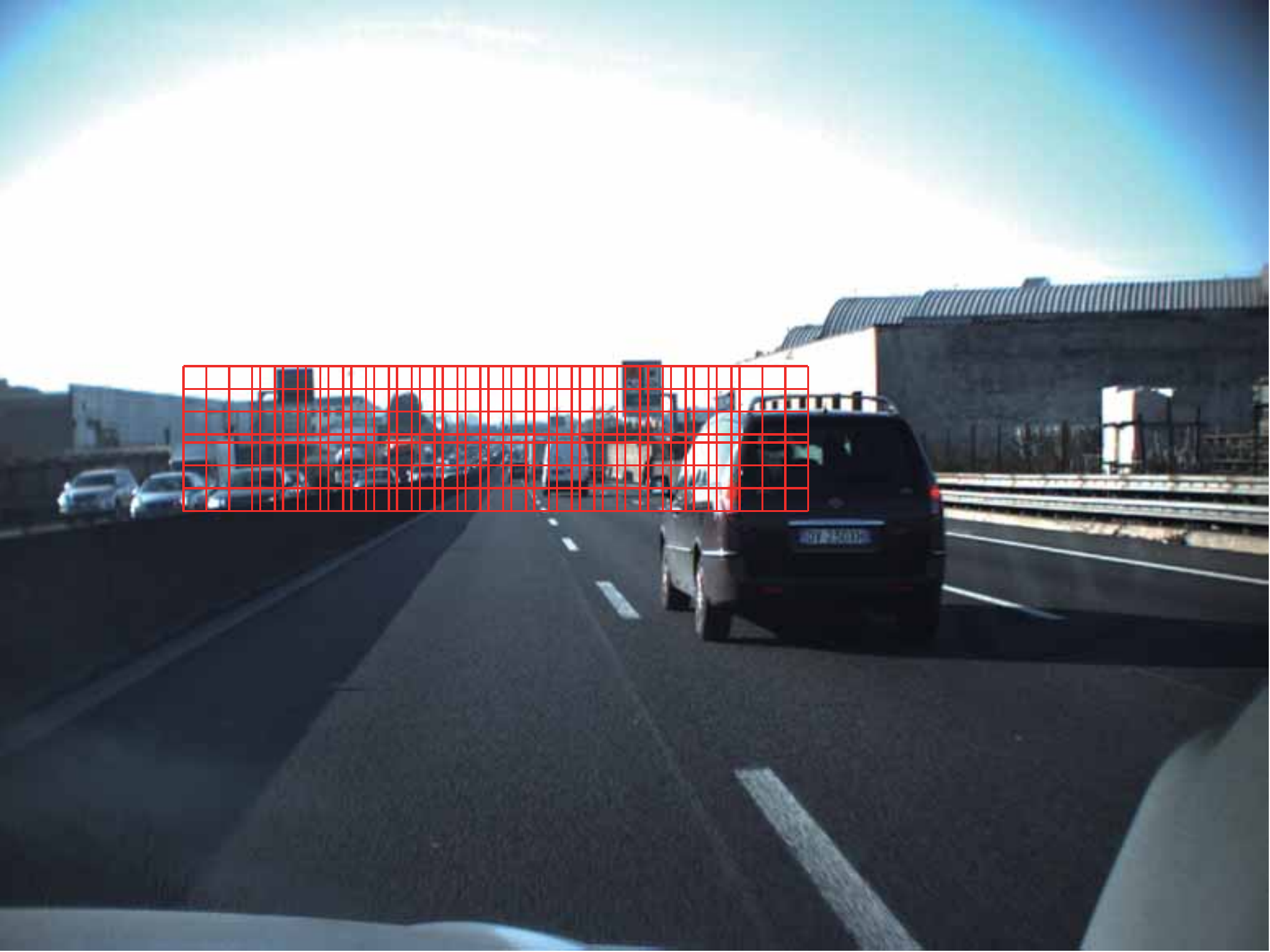}
						}
					\end{minipage}
				}
				\subfloat[]
				{
					\label{fig_Exp_Py08}
					\begin{minipage}[t]{0.24\linewidth}
						\centerline
						{
							\includegraphics[width=1\textwidth]{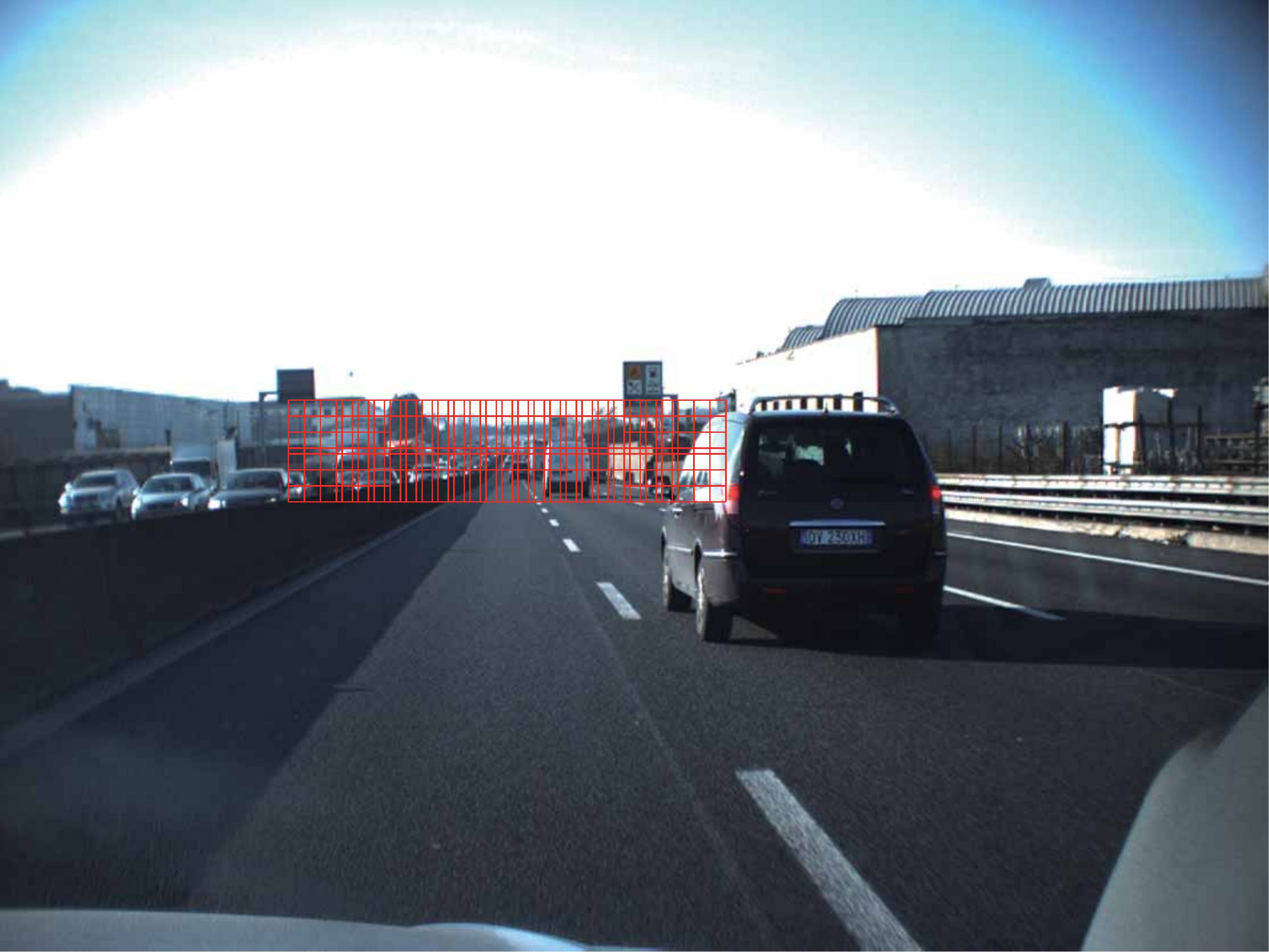}
						}
					\end{minipage}
				}
			}
		\end{minipage}
	}
	\caption{Final sparse windows in each image of the image pyramid.}
	\label{fig_Exp_Pyramid}
\end{figure*}

As shown in Fig.~\ref{fig_Exp_Pyramid}, we illustrate the final sparse windows in each image of the image pyramid using the TME Motorway dataset~\cite{TME12}, which is a challenging and widely-used dataset in vehicle detection. Note that, all of the sparse windows have the same size (\(48 \times 48\) in this work), which means that they can be directly utilized as the input of the deep network. In this way, the sparse windows are avoided to be resized and the efficiency of the approach is ensured.

\section{Experimental Results}
\label{sec_res_dis}

The experiments in this section mainly focus on vehicle detections in traffic scene, which is an important issue in a wide range of applications such as autonomous driving, autonomous navigation and advance driver assistance systems (ADAS)~\cite{Zehang06}. In this work, the performance of our approach is evaluated on a broadly used datasets: TME motorway dataset~\cite{TME12}, which is designed for vehicle detection and localization in challenging traffic scene with various lighting conditions and complex traffic situations~\cite{TME12}.

In the following experiments, the output bounding-boxes of our approach are refined by mean shift and non-maximum suppression, and we use an intersection-over-union (IoU) threshold of 0.7 to determine the correctness of detection. All the experiments in this section are performed on a GTX1080 GPU.

\subsection{Generalized Haar filter based weights vs traditional weights}

As our deep network has two output channels: classification channel and localization channel. We respectively use the classification and localization error of training and test as the performance metrics for the following experiments. The classification error of both training and test is defined as:
\begin{equation}
{
	E{r_{cla}} = \frac{{FP + FN}}{N},
}
\end{equation}
where \(N\) is the total number of training or test samples, \(FP\) and \(FN\) respectively denotes the number of positive and negative samples which are incorrectly classified.

For the localization channel, the error of both training and test is defined as:
\begin{equation}
{
	E{r_{loc}} = \frac{{\sum\limits_{i = 1}^N {{{\left\| {{d_i} - {{\hat d}_i}} \right\|}^2}} }}{{4N}},
}
\end{equation}
where the vector \(d = (d{x_1},d{x_2},d{y_1},d{y_2})\) represents the output of the localization channel, and \(\hat d = (d{{\hat x}_1},d{{\hat x}_2},d{{\hat y}_1},d{{\hat y}_2})\) is the ground truth location vector.

\begin{figure}[t]
	\centering
	{
		\begin{minipage}[t]{0.8\linewidth}
			\centering  
			{
				\subfloat[Classification Error of Training]
				{
					\label{fig_Cla_Train}
					\begin{minipage}[t]{0.48\linewidth}
						\centerline
						{
							\includegraphics[width=1\textwidth]{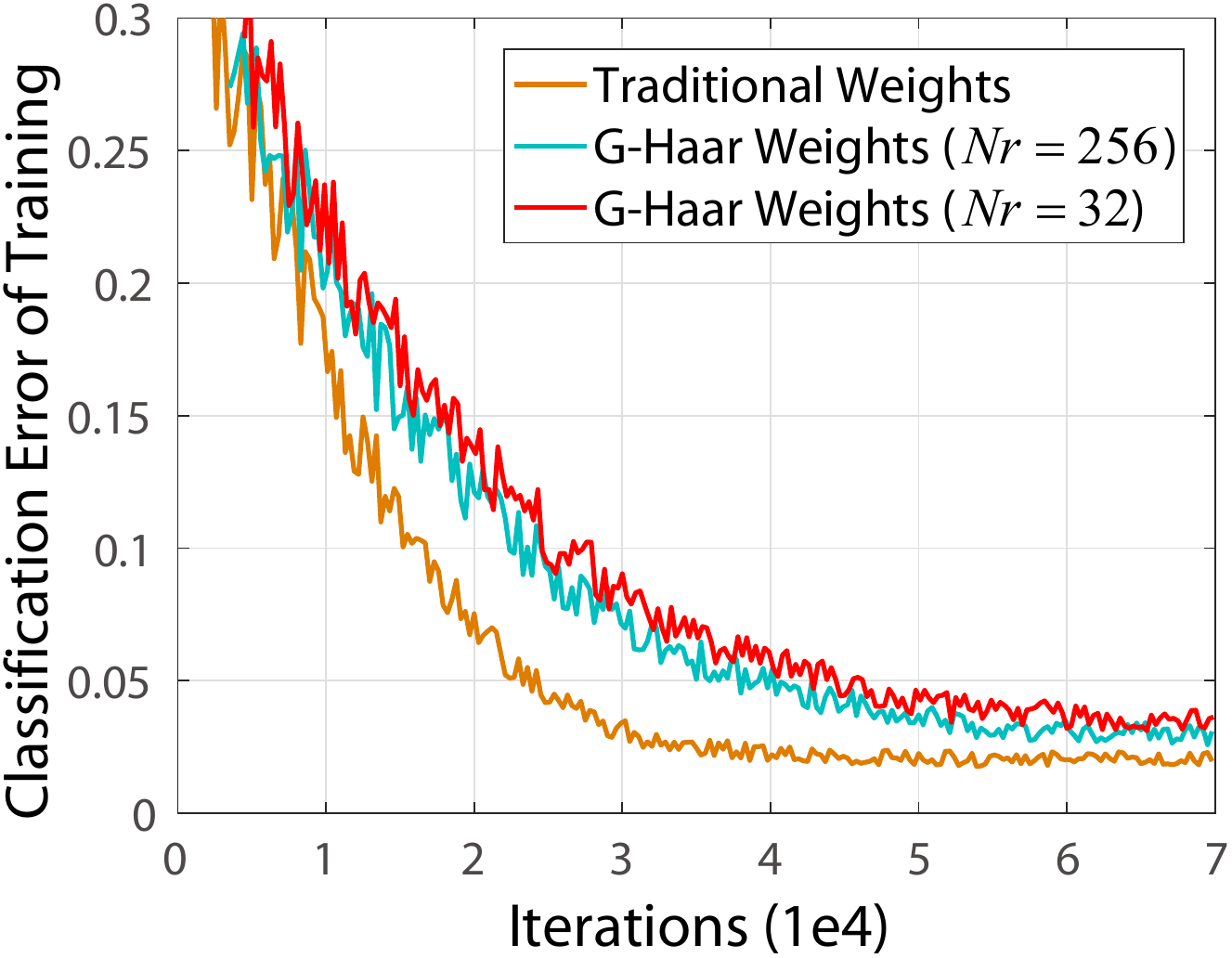}
						}
					\end{minipage}
				}
				\subfloat[Classification Error of Test]
				{
					\label{fig_Cla_Test}
					\begin{minipage}[t]{0.48\linewidth}
						\centerline
						{
							\includegraphics[width=1\textwidth]{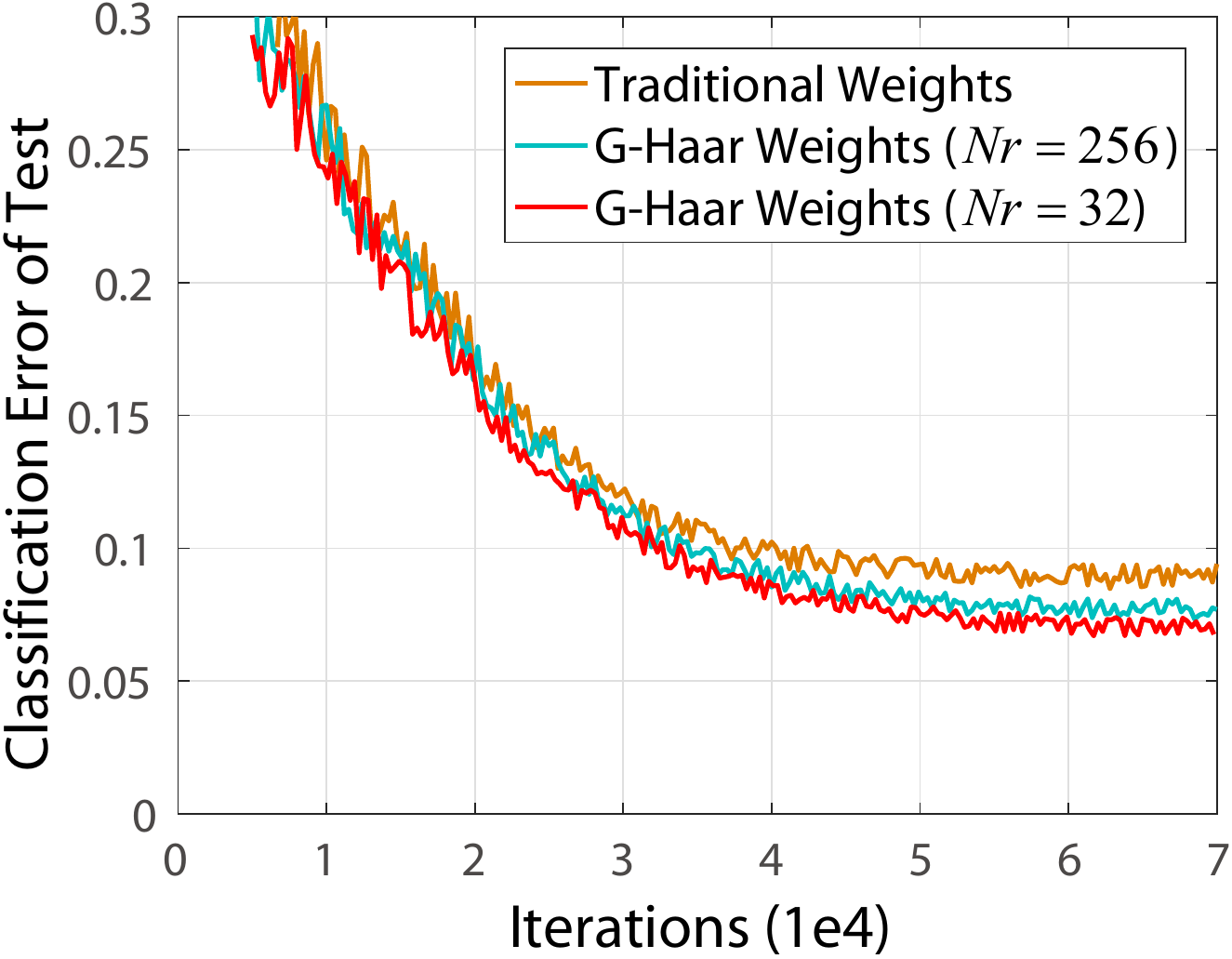}
						}
					\end{minipage}
				}
			}
			\\
			\centering  
			{
				\subfloat[Localization Error of Training]
				{
					\label{fig_Loc_Train}
					\begin{minipage}[t]{0.48\linewidth}
						\centerline
						{
							\includegraphics[width=1\textwidth]{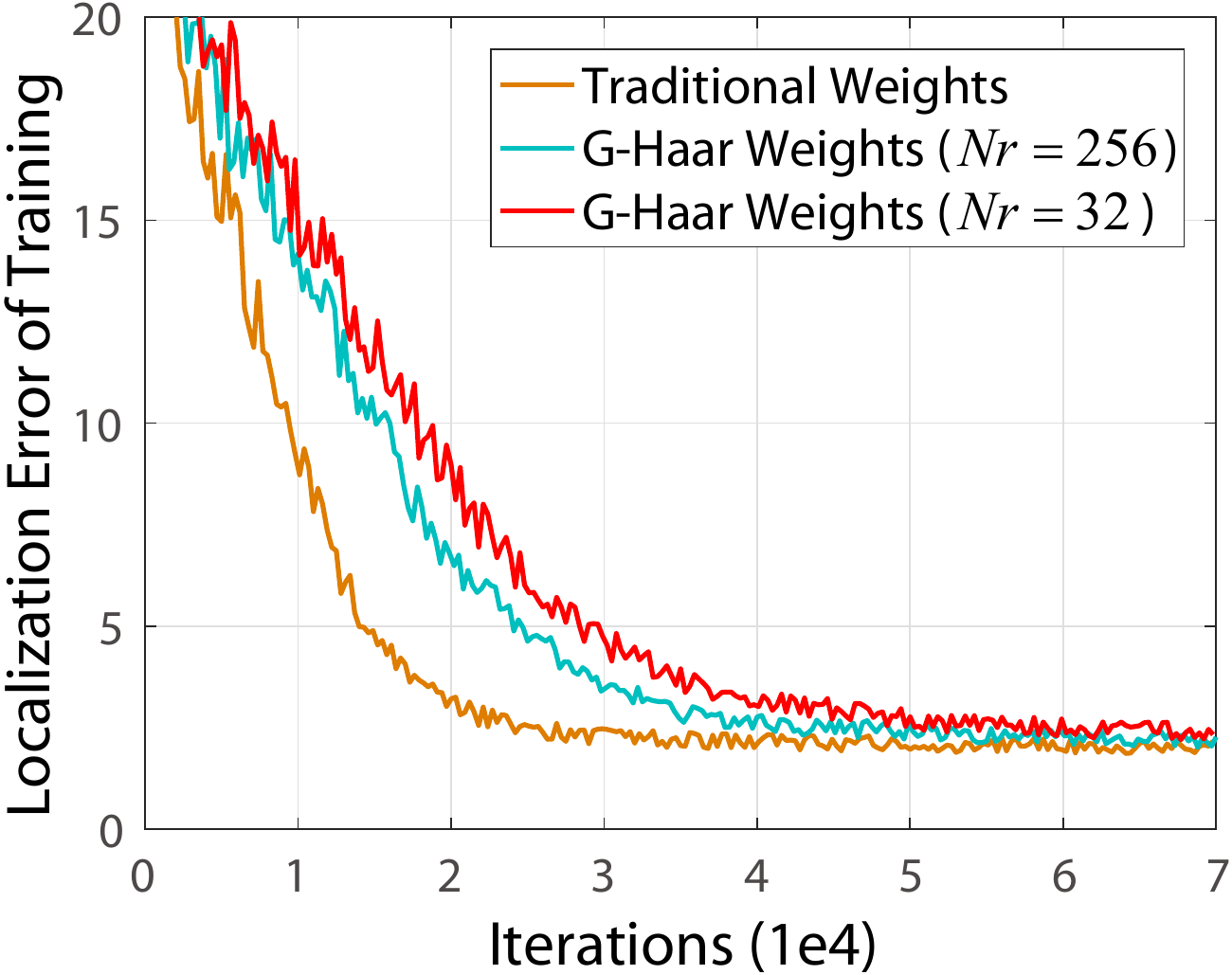}
						}
					\end{minipage}
				}
				\subfloat[Localization Error of Test]
				{
					\label{fig_Loc_Test}
					\begin{minipage}[t]{0.48\linewidth}
						\centerline
						{
							\includegraphics[width=1\textwidth]{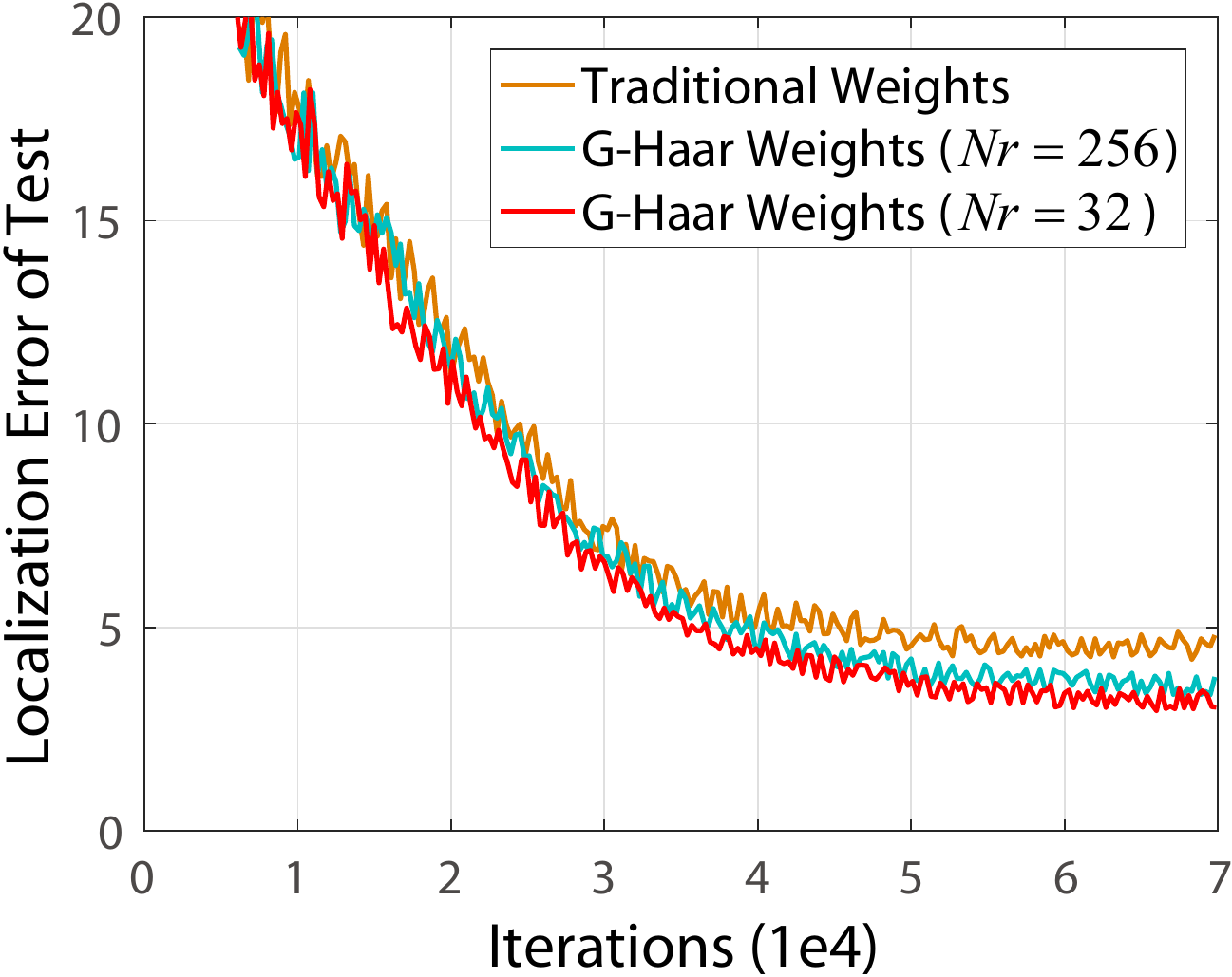}
						}
					\end{minipage}
				}
			}
		\end{minipage}
	}
	\caption{Comparisons of generalized Haar filter based weights (G-Haar weights) and traditional weights on TME motorway dataset.}
	\label{fig_Exp_Tr_vs_Haar}
\end{figure}

Fig.~\ref{fig_Exp_Tr_vs_Haar} illustrates how the errors of classification and localization changed when increasing the number of training iterations. According to this figure, our deep network is able to achieve a high performance when training is convergent. Despite the fact that the deep network with generalized Haar filter based weights (hereafter called G-Haar weights) has a slightly larger training error and needs more training iterations to reach the state of convergence, it is able to produce a less test error in both classification and localization tasks when training is convergent. In other words, the deep network with G-Haar weights has a stronger generalization ability than that with traditional weights. This is owing to the regularization effect of G-Haar weights. Besides, the network employing G-Haar weights with less \(Nr\) (the number of selected filters) needs more training iterations and tend to have stronger generalization ability when training iteration is convergent.

\subsection{Storage and computing resources consumptions}

We investigate the storage and computing resources consumptions in this subsection. The experiment is also performed on TME motorway dataset~\cite{TME12}. In the experiment, the number of selected filters \(Nr\) is 32, and all the weights are stored using single-precision floating-point format (32 bits). In the deep network (see Fig.~\ref{fig_Pipeline_Net}), the size of conv1$\sim$conv5\(\_\)x is \(3 \times 3\), and the rest of convolution kernels have the size of \(1 \times 1\). As the weights of size \(1 \times 1\) consume much less resources than that of size \(3 \times 3\), we only evaluate the dimensions of conv1$\sim$conv5\(\_\)x and their effects on resources consumptions in table~\ref{tbl_source_con}.

As shown in table~\ref{tbl_source_con} (column of Mem.), the networks using G-Haar weights are able to dramatically reduce memory resources (about \(0.8{n^2}\) times, \(n = 3\) in this work). This is due to the fact that only a filter index and a multiplication factor are needed to stored for each G-Haar based convolution kernel, which totally consumes 5 bytes of memory space. By contrast, there are \({n^2}\) weights needed to stored for each traditional convolution kernel of \(n \times n\), thus, \(4{n^2}\) of memory space is required.

Besides, as multipliers are major computing resource consumed by deep networks, we use the number of multiply operations required for each convolution step to measure the consumption of computing resource for the deep networks. As we know, each traditional convolution kernel of \(n \times n\) needs \({n^2}\) multiply operations for each convolution step. Nevertheless, for any convolution kernel using G-Haar weights, each convolution step can be transformed to the form of equation (\ref{eq_convstep}). In this way, only one multiply operation is required for each G-Haar convolution step. Accordingly, as shown in table~\ref{tbl_source_con}  (column of Mul.\({\rm{/}}\)St.), the computing resource consumption of the deep networks using G-Haar weights is only \(1/{n^2}\) of that using traditional weights (\(n = 3\) in this experiment).

In addition, power consumption is directly influenced by storage and computing resources utilizations~\cite{Horowitz2014}. As our deep network with G-Haar weights can markedly reduce storage and computing resources (including memory accesses), it would have a great increase in power-efficiency. This merit is especially meaningful when the deep network implements on embedded systems or mobile devices (such as FPGA and ARM), which are quite sensitive to power consumption.

\begin{table*}[t]
	\centering
	\caption{Storage and computing resources consumptions of our deep network.}
	\label{tbl_source_con}
	\begin{tabular}{c c c c c c c c c c}
		\toprule
		conv1 & conv2 & conv3 & conv4 & conv5\(\_\)x & G-Haar & Mem. & Mul.\({\rm{/}}\)St. & Cla. Err. & Loc. Err. \\
		\midrule
		3\(\times \)64 & 64\(\times \)128 & 128\(\times \)256 & 256\(\times \)256 & 256\(\times \)128 & \(\times\) & 5.97 MB & 9 & 0.147 & 5.19 \\
		3\(\times \)64 & 64\(\times \)256 & 256\(\times \)512 & 512\(\times \)1024 & 1024\(\times \)128 & \(\times\) & 32.13 MB & 9 & 0.135 & 4.78 \\
		3\(\times \)256 & 256\(\times \)512 & 512\(\times \)512 & 512\(\times \)1024 & 1024\(\times \)512 & \(\times\) & 67.74 MB & 9 & 0.092 & 4.21 \\
		3\(\times \)128 & 128\(\times \)256 & 256\(\times \)512 & 512\(\times \)1024 & 1024\(\times \)1024 & \(\times\) & 96.04 MB & 9 & 0.054 & 3.22 \\
		\midrule
		3\(\times \)64 & 64\(\times \)128 & 128\(\times \)256 & 256\(\times \)256 & 256\(\times \)128 & \(\checkmark\) & \ktb{901.96 KB} & \ktb{1} & 0.095 & 4.33 \\
		3\(\times \)64 & 64\(\times \)256 & 256\(\times \)512 & 512\(\times \)1024 & 1024\(\times \)128 & \(\checkmark\) & \ktb{4.51 MB} & \ktb{1} & 0.072 & 3.48 \\
		3\(\times \)256 & 256\(\times \)512 & 512\(\times \)512 & 512\(\times \)1024 & 1024\(\times \)512 &\(\checkmark\) & \ktb{9.58 MB} & \ktb{1} & 0.067 & 2.97 \\
		3\(\times \)128 & 128\(\times \)256 & 256\(\times \)512 & 512\(\times \)1024 & 1024\(\times \)1024 &\(\checkmark\) & \ktb{13.68 MB} & \ktb{1} & 0.042 & 2.32 \\
		\bottomrule
	\end{tabular}
\end{table*}

\subsection{Comparing with state-of-the-art methods} 

\begin{figure}[t]
	\centering
	\includegraphics[width=0.47\textwidth,keepaspectratio]{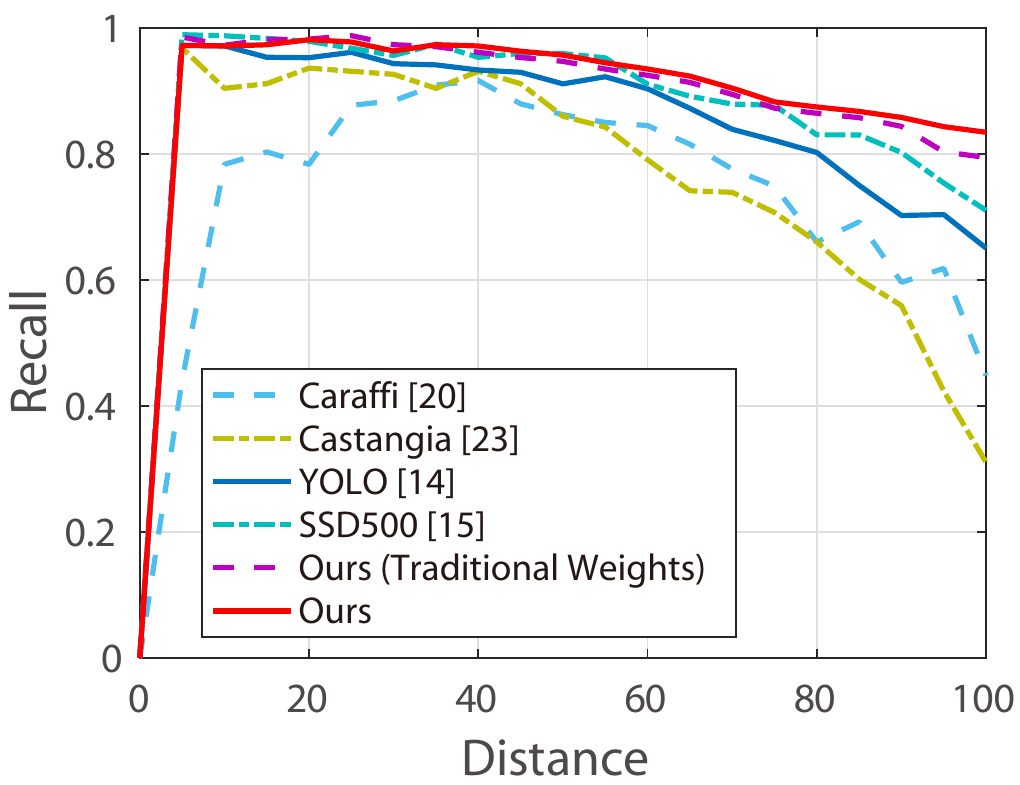}
	\caption{Performance evaluation in different vehicle distances on TME motorway dataset.}
	\label{fig_FIG_TME}
\end{figure}

For further analysis, our proposed approach is evaluated via comparing with some state-of-the-art methods. In the following experiments, ``Ours'' represents our complete method that employs G-Haar weights and the stage of sparse windows generation. ``Ours (Traditional Weights)'' and ``Ours (Without Sparse Windows)'' denote the variants of our approach that utilize traditional weights instead of G-Haar weights and without the stage of sparse windows generation, respectively. The experiment settings of our approaches is the same as that in the row 7 of table~\ref{tbl_source_con}.

As shown in Fig.~\ref{fig_FIG_TME}, thanks to the strong representation of deep convolution neural networks, regression based deep networks (such as YOLO~\cite{YOLO15}, SSD500~\cite{SSD15} and our approach) tend to perform better than traditional hand-craft methods. At the same time, as exhaustive sliding windows can be avoided when using regression based deep networks and convolution can be efficiently computed in GPU, as shown in table~\ref{tbl_Runtime_TME}, regression based deep networks are able to achieve real-time performance in object detection task in traffic scene. Besides, table~\ref{tbl_Runtime_TME} also demonstrates that the stage of sparse windows generation is able to reduce unnecessary computation and improve the efficiency dramatically.

\begin{figure*}[t]
	\centering
	\includegraphics[width=0.9\textwidth,keepaspectratio]{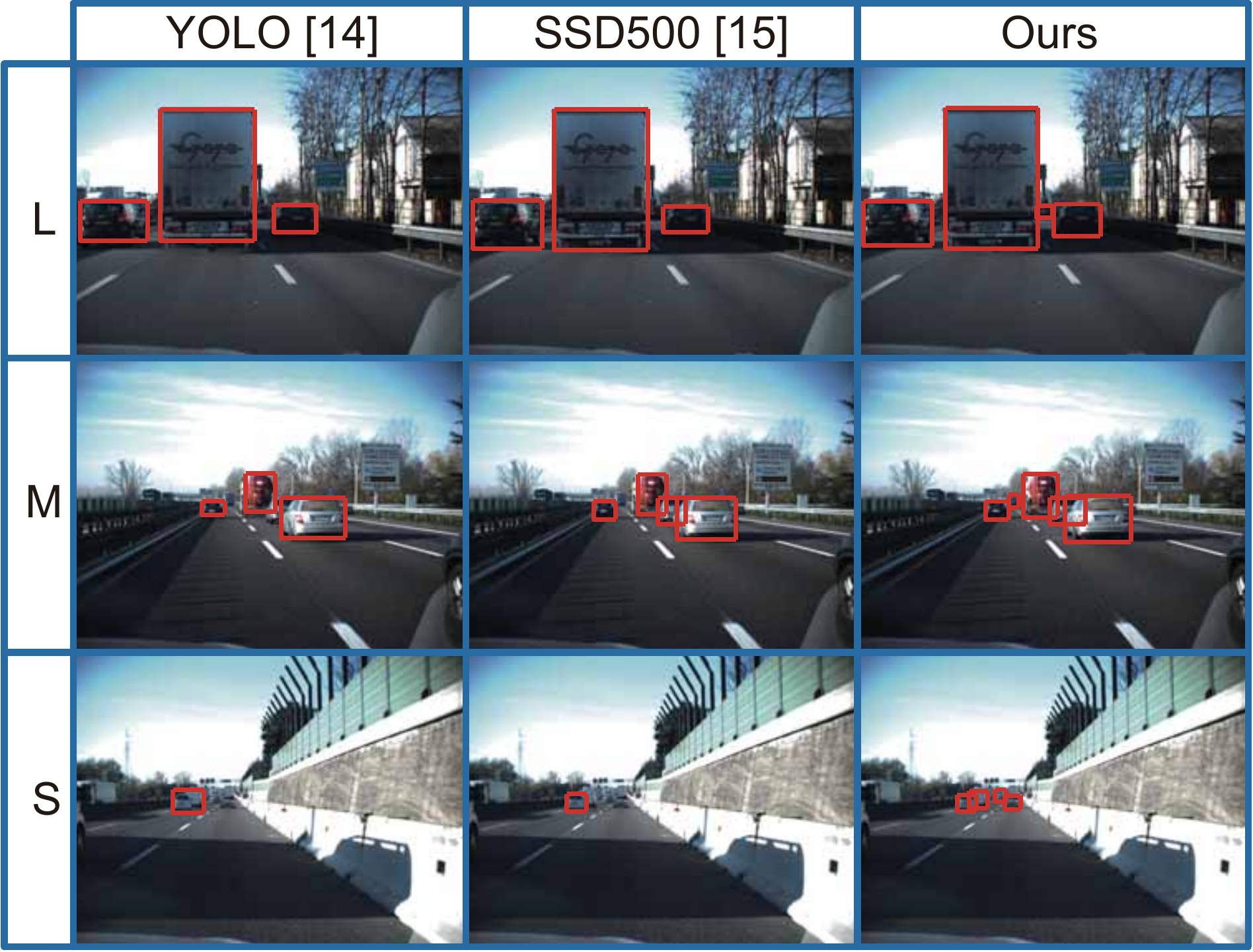}
	\caption{Qualitative results on TME motorway dataset in comparison with two state-of-the-art approaches which are based on deep regression networks. The rows of ``L'', ``M'' and ``S'' respectively mean that the average object sizes are large, middle and small.}
	\label{fig_q_TME}
\end{figure*}

Moreover, quantitative evaluation in Fig.~\ref{fig_FIG_TME} and qualitative results in Fig.~\ref{fig_q_TME} indicate that our proposed method is able to achieve better performance on small object detection compared with other regression based deep networks such as YOLO~\cite{YOLO15} and SSD500~\cite{SSD15} which utilize global regression strategy. SSD500~\cite{SSD15} employs multi-scale feature maps to detect objects in different scales and it perform better than YOLO~\cite{YOLO15}. However, its performance on small objects detection is still unsatisfactory due to the fact that global regression, that is regressing the bounding-box of each object from the whole image, is a much more difficult task compared with our local regression strategy. Beside, input images (e.g. \(1024 \times 768\) in TME motorway dataset~\cite{TME12}) have to be resized to \(500 \times 500\) in SSD500~\cite{SSD15}, which leads to the lacking of resolution of small objects such as vehicles and pedestrians that are far away from camera. Our proposed method decomposes the global regression task into several easier local regression tasks, and detect multi-scale objects from image pyramid. In this way, the resolution of small objects is ensured and thus better performance can be achieve.

\begin{table}[H]
	\centering
	\caption{Runtime analysis on TME motorway dataset.}
	\label{tbl_Runtime_TME}
	\begin{tabular}{c c}
		\toprule
		Method & Average Runtime  \\
		\midrule
		Caraffi~\cite{TME12} & 0.1s \\
		Castangia~\cite{Castangia14} & 0.05s \\
		YOLO~\cite{YOLO15} & 0.022 \\
		SSD500~\cite{SSD15} & 0.043 \\
		Ours (Without Sparse Windows) & 32s \\
		Ours & \textbf{0.025} \\
		\bottomrule
	\end{tabular}
\end{table}

\section{Conclusions and Future Works}
\label{sec_conclu}

In this paper, we presents a novel network system for object detection tasks in traffic scene. In this system, we introduce a local regression strategy for accurate objects detection. Compared with traditional global regression based object detection, the local regression task is easier to handled without the support of the complex or large-scale networks. According to this fact, we handle the local regression tasks by using several tiny deep networks which simultaneously output the bounding boxes, categories and confidence scores of detected objects. In order to satisfy the storage, power and computing source requirements of the platforms for traffic scene applications, the weights of the deep networks are constrained to the form of generalized Haar filter in training phase. Furthermore, to achieve the real-time performance, we introduce the strategy of sparse windows generation to reduce the runtime of our system.

In the experiments, we first evaluate the performance and generalization ability of our generalized Haar filter based weights by comparison with traditional weights. Then the consumptions of storage and computing resources are evaluated. Finally, the effectiveness and efficiency of our approach are validated in comparison with some recently published state-of-the-art methods. Experimental results demonstrate that the proposed approach is proved to be efficient, robust and source saving in challenging traffic scene.

As the proposed approach is suitable for the platforms which have strict limitations on memory, power and computing sources, our future works will focus on implementing the proposed approach (path forward) to FPGA (Field Programmable Gate Array) for advance driver assistance system (ADAS). Owing to the generalized Haar Filter based weights, only one multiply operation is required for each convolution step. This make it possible to construct more parallel pipelines in FPGA where the number of multipliers is limited. Moreover, each local regression task is an independent computation. Thus, all of the local regression tasks can run in parallel in FPGA. In this way, the system can achieve real-time response without extra effort.

\section*{Acknowledgements}
This work is supported by the National Natural Science Foundation of China (Grant No. 61473303).

\bibliographystyle{splncs}

\begin{thebibliography}{10}
	\expandafter\ifx\csname url\endcsname\relax
	\def\url#1{\texttt{#1}}\fi
	\expandafter\ifx\csname urlprefix\endcsname\relax\def\urlprefix{URL }\fi
	\expandafter\ifx\csname href\endcsname\relax
	\def\href#1#2{#2} \def\path#1{#1}\fi
	
	\bibitem{lky15}
	K.~Lu, J.~Li, X.~An, H.~He, Vision sensor-based road detection for field robot
	navigation, Sensors 15(11) (2015) 29594--29617.
	
	\bibitem{ILSVRC}
	{Large Scale Visual Recognition Challenge (ILSVRC)},
	\url{http://www.image-net.org/challenges/LSVRC/}, [Online; accessed Oct. 23,
	2016].
	
	\bibitem{COCO}
	{MS COCO Visual Recognition Challenges}, \url{http://mscoco.org/}, [Online;
	accessed Oct. 23, 2016].
	
	\bibitem{Viola04}
	P.~Viola, M.~J. Jones, Robust real-time face detection, International Journal
	of Computer Vision (IJCV) 57(2) (2004) 137--154.
	
	\bibitem{DPM10}
	P.~F. Felzenszwalb, R.~B. Girshick, D.~McAllester, D.~Ramanan, Object detection
	with discriminatively trained part-based models, IEEE Transactions on Pattern
	Analysis and Machine Intelligence (TPAMI) 32(9) (2010) 1627--1645.
	
	\bibitem{CNN12}
	A.~Krizhevsky, I.~Sutskever, G.~E. Hinton, Imagenet classification with deep
	convolutional neural networks, Advances in Neural Information Processing
	Systems 25 (2012) 1--9.
	
	\bibitem{Wang12}
	T.~Wang, D.~J. Wu, A.~Coates, A.~Y. Ng, End-to-end text recognition with
	convolutional neural networks, in: International Conference on Pattern
	Recognition (ICPR), 2012, pp. 1051--4651.
	
	\bibitem{ResNet15}
	K.~He, X.~Zhang, S.~Ren, J.~Sun, Deep residual learning for image recognition,
	arXiv preprint arXiv:1512.03385.
	
	\bibitem{Fast_RCNN15}
	R.~Girshick, Fast r-cnn, in: IEEE International Conference on Computer Vision
	(ICCV), 2015, pp. 1440--1448.
	
	\bibitem{RCNN12}
	R.~Girshick, J.~Donahue, T.~Darrell, J.~Malik, Rich feature hierarchies for
	accurate object detection and semantic segmentation, in: IEEE International
	Conference on Computer Vision and Pattern Recognition (CVPR), 2014, pp.
	580--587.
	
	\bibitem{SPPNET14}
	K.~He, X.~Zhang, S.~Ren, Spatial pyramid pooling in deep convolutional networks
	for visual recognition, IEEE Transactions on Pattern Analysis and Machine
	Intelligence (TPAMI) 37(9) (2014) 1904--1916.
	
	\bibitem{SelectiveSearch13}
	J.~Uijlings, K.~van~de Sande, T.~Gevers, A.~Smeulders, Selective search for
	object recognition, in: International Journal of Computer Vision (IJCV),
	2013, pp. 154--171.
	
	\bibitem{FasterRCNN16}
	S.~Ren, K.~He, R.~Girshick, J.~Sun, Faster r-cnn: Towards real-time object
	detection with region proposal networks, IEEE Transactions on Pattern
	Analysis and Machine Intelligence (TPAMI).
	
	\bibitem{YOLO15}
	J.~Redmon, S.~Divvala, R.~Girshick, A.~Farhadi, You only look once: Unified,
	real-time object detection, arXiv preprint 1506.02640.
	
	\bibitem{SSD15}
	W.~Liu, D.~Anguelov, D.~Erhan, C.~Szegedy, S.~Reed, C.-Y. Fu, A.~C. Berg, Ssd:
	Single shot multibox detector, arXiv preprint 1512.02325.
	
	\bibitem{DeepCompress}
	S.~Han, H.~Mao, W.~J. Dally, Deep compression: Compressing deep neural networks
	with pruning, trained quantization and huffman coding, Fiber 56(4) (2016)
	3--7.
	
	\bibitem{CompressTracking12}
	K.~Zhang, L.~Zhang, M.-H. Yang, Real-time compressive tracking, in: European
	Conference on Computer Vision (ECCV), 2012, pp. 864--877.
	
	\bibitem{GHaarLike05}
	P.~Doll\'ar, Z.~Tu, H.~Tao, S.~Belongie, Visual tracking with online multiple
	instance learning, in: IEEE International Conference on Computer Vision and
	Pattern Recognition (CVPR), 2009, pp. 983--990.
	
	\bibitem{MinMax11}
	Z.~Yu, An efficient method for solving nonlinear unconstrained min-max problem,
	Master's thesis, Xi'An University of science and technology, Xi'An, China
	(2011).
	
	\bibitem{TME12}
	C.~Caraffi, T.~Voj\'{i}\v{r}, J.~Trefn\'{y}, J.~\v{S}ochman, J.~Matas, A system
	for real-time detection and tracking of vehicles from a single car-mounted
	camera, in: IEEE Intelligent Transportation Systems Conference, 2012, pp.
	975--982.
	
	\bibitem{Zehang06}
	Z.~Sun, G.~Bebis, R.~Miller, Monocular precrash vehicle detection: features and
	classifiers, IEEE Transactions on Image Processing (TIP) 32(9) (2006)
	2019--2034.

	\bibitem{Horowitz2014}
	M.~Horowitz, Computing's energy problem (and what we can do about it), in: IEEE
	International Solid State Circuits Conference, 2014, pp. 10--14.
	
	\bibitem{Castangia14}
	L.~Castangia, P.~Grisleri, P.~Medici, A.~Prioletti, A.~Signifredi, A
	coarse-to-fine vehicle detector running in real-time, in: IEEE Intelligent
	Transportation Systems Conference, 2014, pp. 691--696.
	
\end{thebibliography}

\end{document}